\documentclass[12]{congress}
\usepackage{epsfig}
\usepackage{overcite}
\usepackage{isolatin1}
\usepackage{graphics}
\title{Using Domain Knowledge in Evolutionary System Identification}
\vspace{18pt}

\address{\parbox{7.6cm}{ \bf\small Marc Schoenauer\textsuperscript{$\star$} \\
                         \it\footnotesize Projet Fractales \\
                            INRIA Rocauencourt, B.P. 105\\
                            78153 LE CHESNAY Cedex, France \\
                            Marc.Schoenauer@inria.fr \\
                            http://www-rocq.inria.fr/fractales/Staff/Schoenauer}
  \hfill \parbox{6.8cm}{\bf\small Michèle Sebag \\
                        \it\footnotesize Laboratoire de Mécanique des Solides \\
                            Ecole polytechnique \\
                            91128 Palaiseau Cedex, France \\
                            Michele.Sebag@polytechnique.fr, \\
                            http://www.eeaax.polytechnique.fr/michele} }

\heading{M. Schoenauer and M. Sebag/Evolutionary System Identification}

\abstract{Two example of Evolutionary System Identification are
  presented to highlight the importance of incorportaing Domain
  Knowledge: the discovery of an analytical indentation law in Structural
  Mechanics using constrained Genetic Programming, and the
  identification of the repartition of underground velocities in
  Seimsi Prospection. Critical issues for sucessful ESI are discussed
  in the light of these results.
}


\begin{document}

\section{EVOLUTIONARY SYSTEM IDENTIFICATION}
In the recent decades, modelisation and simulation of physical
phenomena has been  widely applied to almost all areas of engineering:
from given experimental and initial conditions, the outcome of the
physical system are computed, using some (generally simplified) {\em
  model} of the underlying phenomena.

Two slightly different situations pertain to System Identification:
When the model is totally unknown, the goal is to find the
relationship between the input and the output of the whole system, as
is the case in the problem presented in section \ref{meca}.
In some other domains, the model itself relies on some sub-model:
the  {\em   microscopic} behavioral law in Solid Mechanics, the {\em
  local} command in control problems, the velocity repartition in the
underground (section\ref{petrole}). The goal is then to discover the
sub-model that gives correct predictions of the output when
used in the higher-level model.

Nevertheless, in both cases, the unknown is a function,
and some common issues arise. The most important one is of course the
choice of the search space, and some trade-off has to be made:
parameterized functional spaces (e.g. spline functions) allow one to
use deterministic optimization methods -- though the resulting
optimization problem generally is highly irregular and multi-modal; on
the other hand, unstructured search spaces require stochastic methods
-- and this is explains the recent progresses in Evolutionary System
Identification. 

But Evolutionary Computation  always faces a dilemma. On the one hand,
its flexibility to handle any type of
search space opens up huge possibilities, such as
searching spaces of graphs, of variable lengths lists, \ldots ~
Such weird search spaces, being supersets of classical
parameterized search spaces, certainly contain better solutions for the
problem at hand.
On the other hand, the size of these search spaces can hinder the
search, and the Evolutionary Algorithm is likely 
to get stuck in local optima, or to explore only a limited part of the
whole search space.

Possible answers to that critical choice is problem specific, and 
this paper addresses will try to enlighten 
this issue through two test cases in the area of System Identification
Domain knowledge can be used to restrict the search space,
as section \ref{meca} will demonstrate by introducing a constrained
version of Genetic Programming 
based on Context-Free Grammars in order to cope with dimensional
analysis in Structural Mechanics system identification. But even
after choosing a smart representation, the search might be hindered by
a bad choice of 
fitness function: section \ref{petrole} will present an example of
identification of the underground in geophysical
seismic prospection. In that context, domain knowledge implicitly
relies on some common sense assumptions that have to be taken into
account explicitly in the Evolutionary Algorithm.
General issues related to the introduction of domain knowledge in
Evolutionary System Identification will then be 
discussed in the light of these results  in the final section.

\section{DIMENSIONAL CONSTRAINTS IN STRUCTURAL  MECHANICS}
\label{meca}
This section focuses on dimensional consistency,
a most common background knowledge in scientific domains:
given the units associated to
the problem variables, the target solution should be a well-formed
dimensioned expression (one should not add meters and seconds).
All work presented in this section is joint work of the second author
with A. Ratle (ISAT, Nevers, France)
 \cite{Ratle-Sebag:PPSN2000,Ratle-Sebag:EA01}.

\subsection{The mechanical problem and dimensionality constraints}
\label{meca_problem}
The test-case presented here is a simplified real-world application,
macro-me\-cha\-nical modeling from indentation tests.
Indentation tests proceed by pressing a hard indenter 
(of conical or tetrahedral shape) against  the surface of the
material to be tested out.  The experimenter records the reaction
force along time and displacement. The target solution (materials 
behavioral law) expresses 
force at time $t$ as an analytical function of the materials
parameters, and the displacement and its derivative at time $t$.

The state of the art in Mechanics provides such analytical expressions for 
simple constitutive laws only 
For known
materials with complex constitutive laws, 
the indentation law is obtained from computationally 
heavy simulations (a few hours on an HP350 workstation). For ill-known 
materials, no indentation law at all is provided. In the two latter
cases, behavioral law modeling is a challenge to non parametric
model identification.

A standard evolutionary approach to non-parametric identification
is Genetic Programming \cite{Banzhaf:book}. 
For the sake of procedural simplicity, canonical GP strongly
relies on the closure hypothesis (replacing any  subtree by
another one results in a viable individual). However,  
the set of dimensionally
consistent models is a only tiny fraction of the model set.
 
Enforcing dimensionality consistency amounts to constraint GP
search. One straightforward way to add constraints to any
optimization method is to penalize the
individuals that violate the constraints. Along this line, a
penalisation-based approach has been proposed for dimensionally-aware 
identification \cite{Keijzer99}. 
However, restricting the search to the feasible space is, when 
possible, a more efficient approach in Evolutionary 
Computation \cite{Marc:Zb:PPSN96}.

\subsection{Grammar-guided GP}

An elegant approach for getting rid of syntactic constraints in GP
\cite{Gruau96}, is to combine GP with  
Backus Naur Form (BNF) grammars. 
BNF grammars describe the admissible constructs of a language through 
a 4-tuple $\{S,N,T,P\}$: $S$ denotes the start symbol, $N$ the set of
non-terminals, $T$ the set of terminals, and $P$ the production
rules. 
Any expression is built up from the start symbol. Non-terminals
are rewritten into one of their derivations 
as given by the production rules, until 
the expression contains terminals only.

Interestingly, the construction of any admissible expression through
the application of  derivation rules, can itself be represented as 
a tree termed {\em derivation tree}. {\em Grammar Guided GP} (G3P)
proceeds by exploring the space of derivation trees. 
The crossover and mutation operators 
are modified to produce valid derivation trees from valid parents, 
in the spirit of strongly typed GP \cite{Montana95}. 

\smallskip
\noindent
\textbf{Generation of Dimensional Grammars}\\
In order to apply Grammar-guided GP to dimensionally consistent
identification, a BNF grammar encoding dimensional consistency 
can be automatically generated, under the restriction that a finite 
number of units are considered.

In the indentation-based modeling application
(section \ref{meca_problem}), the 
elementary domain units correspond to  mass, length and time.
Every compound unit is represented by its exponents w.r.t. to the basic
units \cite{Keijzer99}. Only integer exponents are considered; for
instance, the Newton unit ($kg\times  m / s^2$) is represented as the 
triplet $(1,1,-2)$. In what follows, 
the exponent of each compound unit  is in a given integer range $[-2,2]$.

To each allowed compound unit $(i,j,k)$ is associated a non terminal
$<N_{i,j,k}>$.
The associated production rules describes all ways of constructing an 
expression with type $(i,j,k)$: by adding or subtracting two expressions
of type $(i,j,k)$; by multiplying two expressions with types $(l,m,n)$
and $(o,p,q)$  such
that $l+o=i$, $m+p=j$, $n+q=k$; and so forth.
Finally, the derivation rule associated to the start symbol gives the unit of
the target expression (in Newton):
$S ~:= ~<N_{1,1,-2}>$

\smallskip
\noindent
\textbf{Initialization}
\label{meca_init}

An unexpected difficulty arose during the initialization step.
In a non-toy grammar such as above, 
the fraction of terminal derivations is so small that uniform 
initialization tends to grow {\em very} deep
and long trees \cite{Ryan98}. Adding an upper bound on the depth does
not help much: 
much time is wasted in generating overly long trees, and 
rejecting them.
What is worse, the initial population is ultimately composed of 
poorly diversified individuals; and final results are poor as evolution
hardly recovers from this initial handicap.

A first attempt to solve this difficulty was to select terminals with higher
probabilities.
Unfortunately, the adjustment of these probabilities proved time-consuming, 
without significantly improving the diversity in the initial population.

The initialization problem was finally addressed in the spirit of constraint
propagation: a depth index is attached to each symbol and 
derivation in the grammar, and records the depth of the smallest tree
needed to rewrite the symbol into a terminal expression (the index is 
recursively computed beforehand). 
From the depth index, one can determine whether a given derivation
rule is admissible in order to rewrite a non-terminal symbol 
in the current expression, i.e. compatible with the total tree depth.
Constrained initialization then proceeds by uniformly sampling one among
the admissible derivation rules.

\subsection{Application}\label{resuinden}
\label{dimension}
The scalability of G3P has been investigated on the 
problem of materials behavioral law modeling from indentation tests
(section \ref{meca_problem}).
Although the grammar size is 
exponential in the allowed range ($|D| = 5^3 = 125$ non-terminal
symbols are considered), its use entails no   
computational overhead compared to a procedural 
dimensional consistency check.

One claimed advantage of using background 
knowledge, e.g. reducing the size of the search space, can be seen by
actually computing the sizes of the search spaces for given maximum
depths: unconstrained GP gives figures like 8.20125e+5,  5.54899e+14
and 3.47438e+23 for depths of respectively 10, 22 and 34, to be
compared to  24, 5.53136e+7 and 1.02064e+14 for G3P. 
Though the size of the search space explored by G3P still 
grows has exponentially with the maximum depth, 
it demonstrates an exponential gain over GP.


Furthermore, and as could have been expected from statistical learning
theory,  the reduction of the search space does improve 
the search efficiency: the results  obtained with 
dimensional grammars always supersede those obtained with untyped 
grammars by an average of 6 standard deviations
\cite{Ratle-Sebag:PPSN2000,Ratle-Sebag:EA01}.

\section{SEISMIC UNDERGROUND PROSPECTION}
\label{petrole}
One of the most challenging problems of the last twenty years in
petroleum prospection  is the determination of the structure of the underground
from data from seismic geophysical experiments. The goal of the inverse
problem in seismic reflection is to identify the velocity distribution
in the underground from recorded reflection profiles of acoustic
waves.
All work presented in this section is joint work of the first author
with F. Mansanné (Université de Pau et des Pays de l'Adour) in
collaboration with the IFP (French Pretoleum Institute). 
All details can be found (in French) in F. Mansanné's PhD, or 
in other publications \cite{IFP:ICEC98,ISMIS99,Mansanne:SoftComputing01}

\subsection{The geophysical problem} 

A seismic experiment starts with an artificial explosion at some point
near the surface. The acoustic waves propagate through the underground
medium, eventually being reflected by multiple interfaces between
different media. 
The reflected waves are measured at some points of
the surface by some receptors recording the pressure variations 
 along time, called {\em seismograms}.
The identification problem is to identify the repartition
of the velocities in the underground domain  from the seismograms.

Such geophysical inverse problem results in a  highly nonlinear,
irregular and multi-modal objective function.
Consequently, local optimization approaches, like steepest
descent or conjugate gradient, are prone to be trapped in local
optima. Hence Evolutionary Algorithms have been long used to tackle
this problem
\cite{Stoffa-Sen-91,Jin:Madariaga:96,Docherty}.

\subsection{Representations for underground identification}
\label{petrole_representation}

However, all the above-mentioned works are based on a
parametric representation of the underground: either a prescribed
layout of the velocities is assumed (the so-called {\em blocky} model,
where velocities are assumed piecewise constant), and the only
unknowns are the velocity values themselves
\cite{Stoffa-Sen-91,Jin:Madariaga:96}; or some global
approximation technique is used (e.g. splines \cite{Docherty}) and the
parameters to identify are the coefficients of that approximation.

Assuming a {\em blocky} model, and because the fitness computation had
to rely on some discretization of the underground domain, one could
also think of a parametric representation attaching one velocity value to
each element of a given mesh. However, such a representation does not
scale up with the mesh size, and/or when going to 3-dimensional
problems. Thus, as discussed in \cite{Hamda:ACDM00} in a
slightly different context, 
a non-parametric variable length representation based on \textbf{Voronoi
  Diagrams} was chosen:
The genotype is a (variable length) list of points (the Voronoi
sites), in which each site is attached a  velocity. The
corresponding Voronoi diagram 
is constructed, and each Voronoi cell is given the velocity of the
corresponding site.

The variation operators are \cite{Hamda:ACDM00}:
a {\bf geometrical} crossover: a random line is drawn across
  both parents, and the Voronoi sites on one side of that line are
  exchanged, several mutation operators based on Gaussian mutations of
  the real-valued parameters, and mutations by addition or
  destruction of Voronoi sites.

\subsection{The fitness functions}
\label{fitnesses}
The first idea is
to use a simulation of the wave equation for the direct problem, 
and to compare the
simulated results to the experimental ones.
This  standard approach has been used in most previous works 
\cite{Stoffa-Sen-91,IFP:ICEC98}. 
Thus, the identification
problem is turned into the minimization of some least square
error function (the {\em LS fitness}). 

An alternative approach proposed by the domain experts consists in
retrieving the velocity background by using the focusing property of
pre-stack depth migration to update the velocity model 
\cite{Varela94,Docherty}.
In an image gather, each trace represents a migrated
image of the subsurface at the same horizontal position.  The
{\em Semblance fitness} relies on the fact that reflection
events in an 
image gather are horizontally aligned if the underground velocity
model is correct. To measure the
horizontal alignment of the reflection events in an image gather, the
criterion first proposed in \cite{Taner69}, and applied with
success in \cite{Jin:Madariaga:96} to 1D seismic profile from the North
Sea, has been used.

The main advantage of the migration velocity analysis methods is that
they are well understood by geologist experts, and are one order of
magnitude faster in terms 
of computing time compared to solving the wave equation \cite{Stoffa-Sen-91}.

\subsection{Results}
\label{petrole_results}
Due to its lower computational cost, and because the domain experts
considered it a more robust criterion than the least square comparison
of simulated and experimental seismograms, first experiments on
realistic models of the underground (the IFP model {\em Picrocol}) used the
the Semblance fitness \ldots with disastrous
results \cite{ISMIS99}: the experimental seismograms were actually
simulated on known synthetic models of the underground, so the
solution was known. However, some totally unrealistic solutions
emerged, that had better Semblance than the actual solution.
Of course, such parasite solutions would never have been retained by
even first grade students in Geophysics. But there was nowhere in the
modelisation of the problem where such common sense argument could be
added (e.g. sandwiches of low velocity between two layers of
very high velocity).

On the other hand, using the LS fitness based on a numerical
simulation of the wave equation did not exhibit this defect on coarse
discretization \cite{IFP:ICEC98}, but proved far too costly for
more realistic discretizations. A compromise was finally set up and
successfully used: alternating both
methods, i.e. using the Semblance fitness during a few (5-10)
generations, then the LS fitness during some (2-5) generations, proved
both robust (weird solutions with good Semblance were eliminated by
the few generations using LS fitness) and not to costly (in addition,
an increasing number of shots were taken into account - one wave
equation must be solved for each shot).

The best results for some subset of the Picrocol model discretized
according to a $80 \times 70$ mesh reached an average relative error
of around 12\%, less than 5\% in 3/4 of the domain, but still required
200h of Silicon O2 computer for 20000 evaluations of each
fitness. These results were considered very encouraging by the domain experts.

\section{Discussion and conclusion}
\label{discussion}
The choice of a representation very often is dictated by \ldots the
target optimization method. Indeed, the only optimization methods that
have been around for a long time in the Applied Maths field are
the standard deterministic numerical optimization methods that are
only defined on {\em parametric search spaces} (i.e. looking ofr vectors of
real-numbers). Hence, the first thought of a numerical engineer facing
an optimization problem is to transform that problem into a parametric
optimization problem. However, such transformation might imply some
restrictions on the search space, making optimal solutions of the
original problem out of reach. A good example is given on the
geophysical test-case (section \ref{petrole_representation}) where
splines, or fixed-complexity blocky models had been used exclusively
because the resulting problem amounted to parametric optimization. 
It is interesting to note that even the first Evolutionary approaches
did not question the representation 

On the opposite, choosing a very general representation that adds
the least possible limitations on the problem might result in a huge
search space, where indeed some very good solutions lie, by in which
the exploration might be very difficult, resulting in sub-optimal
solutions too. It it then necessary to restrict the search to some
particular regions of the search space. However, there are generally
many regions of those huge general search space that common sense
considerations could easily prune -- but EAs lack common sense!

On the seismic prospection problem (section
\ref{petrole_representation}), the search space should be further
reduced: for instance, experts very well
know that there are not large jumps of velocity values between the
different blocks, that deep parts of the underground cannot have small
velocities, etc But such considerations are very difficult to take
into account. On the opposite, in the Mechanical problem of section
\ref{meca}, expert knowledge (dimensional analysis) is the basis of
the restriction of the huge GP search space.\\

Dual to the choice of a representation is the choice of variation
operators (mutation, crossover and the like). It has been argued that
a proper choice of representation could naturally lead to good
operators \cite{Surry:PPSN96} -- and indeed such approach certainly
avoids using  useless (i.e. meaningless for the problem at hand)
operators. However, domain knowledge can also clearly help improving
such representation-independent operators. For instance, the crossover
operator based on simple geometrical considerations used
for the Voronoi representation described in section
\ref{petrole_representation} does outperform the blind
exchange of Voronoi sites \cite{KS_EA95}. \\

Similarly, the initialization is another part of an Evolutionary
Algorithm where domain knowledge can usefully step in. Initialization
is very often taken for granted in Evolutionary Algorithms, e.g. a
common advice is to perform a uniform sampling of the search space. 

First, even when such a uniform distribution does exist on the search space,
some problem specific issues might lead to non-uniform initialization
\cite{Kallel:ICGA97}. 
But there are many cases where nothing like a uniform distributions on the
search space actually exists. 
Hence some {\em ad hoc} procedures have to be designed, and
domain knowledge is mandatory there, too. A very simple example is the
case of unbounded real parameters, where some distribution has to be
arbitrarily chosen (e.g Gaussian with given standard deviation,
uniform on a bounded interval, \ldots). 
This is of course even more difficult in the case of variable length
representations. In standard GP, for instance, the now-standard ramp
half-and-half procedure \cite{Banzhaf:book} took some years to design.
Finally, restricting the search space as is done in the Structural
Mechanics example also implies modifying the initialization procedure,
and that might prove quite difficult (see section \ref{meca_init}).\\

Last but not least, the choice of the fitness function must be
addressed with care -- and this might be surprising
to many researchers: the objective function is generally the starting
point and is very often never even discussed. But that starting point
very often is already the result of some simplification, or some
choice from the modelizing engineer -- and he might have made the
wrong choice on some important decision. For instance, 
though the least square comparison is generally preferred, the maximum
difference is another possibility in the simple case
where the objective function is built on the aggregation of 
different fitness cases. Moreover, as demonstrated in section
\ref{petrole}, some widely accepted objective functions might in fact
assume some common sense rules that do not exist -- and will be very
difficult to add -- in the Evolutionary Algorithm that will actually
solve the problem.\\

In summary, we believe that Evolutionary Algorithms will have a large
impact in  
System Identification -- similarly to that of
Artificial Creativity in the area of Design \cite{Bentley:Corne:2001}.
However, there are many obstacles on the road to success, and this
paper has tried to highlight some of them, hoping that this will make
future evolutionary engineers ask themselves the important questions
before entering the optimization loop.

{\small
\bibliography{../../Bib/LA_TOTALE}
}
\end{document}